\documentclass[nai,crcready]{iosart2x}

\usepackage{appendix}
\usepackage{microtype}
\graphicspath{{img/}}

\newcommand{\R}{\mathbb{R}}
\DeclareMathOperator*{\argmax}{arg\,max}

\pubyear{0000}
\volume{0}
\firstpage{1}
\lastpage{1}

\begin{document}
\begin{frontmatter}

\title{From Neural Activations to Concepts: A Survey on Explaining Concepts in Neural Networks}
\runtitle{From Neural Activations to Concepts: A Survey on Explaining Concepts in Neural Networks}

\begin{aug}
    \author{\inits{J. H.}\fnms{Jae Hee} \snm{Lee}\ead[label=e1]{jae.hee.lee@uni-hamburg.de}%
      \thanks{Corresponding author. \printead{e1}.}} \author{\inits{S.}\fnms{Sergio} \snm{Lanza}\ead[label=e2]{sergio.lanza@uni-hamburg.de}} \author{\inits{S.}\fnms{Stefan} \snm{Wermter}\ead[label=e3]{stefan.wermter@uni-hamburg.de}} \address{Knowledge Technology Group, Department of Informatics, \orgname{University of Hamburg}, \cny{Germany}\printead[presep={\\}]{e1,e2,e3}}
\end{aug}


\begin{abstract}
    In this paper, we review recent approaches for explaining \emph{concepts} in neural networks. Concepts can act as a natural link between learning and reasoning: once the concepts are identified that a neural learning system uses, one can integrate those concepts with a reasoning system for inference or use a reasoning system to act upon them to improve or enhance the learning system. On the other hand, knowledge can not only be extracted from neural networks but concept knowledge can also be inserted into neural network architectures. Since integrating learning and reasoning is at the core of neuro-symbolic AI, the insights gained from this survey can serve as an important step towards realizing neuro-symbolic AI based on explainable concepts.


\end{abstract}

\begin{keyword}
    \kwd{Explainable artificial intelligence, concept explanation, neuro-symbolic integration}
\end{keyword}

\end{frontmatter}

\section{Introduction}
\label{sec:introduction}

In recent years, neural networks have been successful in tasks that were regarded to require human-level intelligence, such as understanding and generating images and texts, performing dialogues, and controlling robots to follow instructions~\cite{openai_gpt-4_2023,ramesh_zero-shot_2021,jiang_vima_2023}. 
However, their decision-making is often \emph{not} explainable, which undermines user trust and negatively impacts their usage in sensitive or critical domains, such as automation, law, and medicine. One way to overcome this limitation is by making neural networks explainable, e.g., by designing them to generate explanations or by using a \emph{post-hoc} explanation method that analyzes the behavior of a neural network after it has been trained.



This paper reviews explainable artificial intelligence (XAI) methods with a focus on explaining how neural networks learn \emph{concepts}, as concepts can act as primitives for building complex rules, presenting themselves as a natural link between learning and reasoning~\cite{schockaert_modelling_2022}, which is at the core of neuro-symbolic AI~\cite{hitzler_compendium_2023,wermter_hybrid_1989,WPA99,mcgarry_rule-extraction_1999,lee_chapter_2023}. On the one hand, identifying the concepts that a neural network uses for a given input can inform the user about what information the network is using to generate its output \cite{bau_network_2017,koh_concept_2020,kim_interpretability_2018,mu_compositional_2020,hernandez_natural_2022,wermter_knowledge_2000}. 
Combined with an approach to extract all relevant concepts and their (causal) relationships, one could generate explanations in logical or natural language that faithfully reflects the decision procedure of the network. On the other hand, the identified concepts can help a symbolic reasoner intervene in the neural network such that debugging the network becomes possible by modification of the concepts~\cite{bau_gan_2018,koh_concept_2020,abid_meaningfully_2022,meng_locating_2022}. 

Some XAI surveys have been published in recent years~\cite{dwivedi_explainable_2023,ibrahim_explainable_2023,madsen_post-hoc_2022,ras_explainable_2022,li_interpretable_2022,samek_explaining_2021,sado_explainable_2023}. However, almost all of them are mainly concerned with the use of saliency maps to highlight important input features. Only a few surveys include concept explanation as a way to explain neural networks. A recent survey in this vein is by Casper et al.~\cite{casper_toward_2023}, which discusses a broad range of approaches to explaining the internals of neural networks. However, due to its broader scope, the survey does not provide detailed descriptions of methods for explaining concepts and misses recent advances in the field. The surveys by Schwalbe~\cite{schwalbe_concept_2022} and Sajjad et al.~\cite{sajjad_neuron-level_2022}, on the other hand, are dedicated to specific kinds of concept explanation methods with a focus on either vision~\cite{schwalbe_concept_2022} or natural language processing~\cite{sajjad_neuron-level_2022} and are, therefore, limited in scope, failing to analyze the two areas together.

We categorize concept explanation approaches and structure this survey based on whether they explain concepts at the level of individual neurons (\autoref{sec:expl-activ-neur}) or at the level of layers (\autoref{sec:explain-activ-layer}). The last section summarizes this survey with open questions.



\section{Neuron-Level Explanations}
\label{sec:expl-activ-neur}

The smallest entity in a neural network that can represent a concept is a \emph{neuron}~\cite{sajjad_neuron-level_2022}, which could be---in a broader sense---also a \emph{unit} or a \emph{filter} in a convolutional neural network~\cite{bau_network_2017}. In this section, we survey approaches that explain, in a post-hoc manner, concepts that a neuron of a pre-trained neural network represents, either by comparing the \emph{similarity} between a concept and the activation of the neuron (see \autoref{sec:similarity-based}) or by detecting the \emph{causal relationship} between a concept and the activation of the neuron (see \autoref{sec:causality-based}).



\subsection{Using Similarities between Concepts and Activations}
\label{sec:similarity-based}

\begin{figure}[t]
    \centering%
    \includegraphics[scale=0.7]{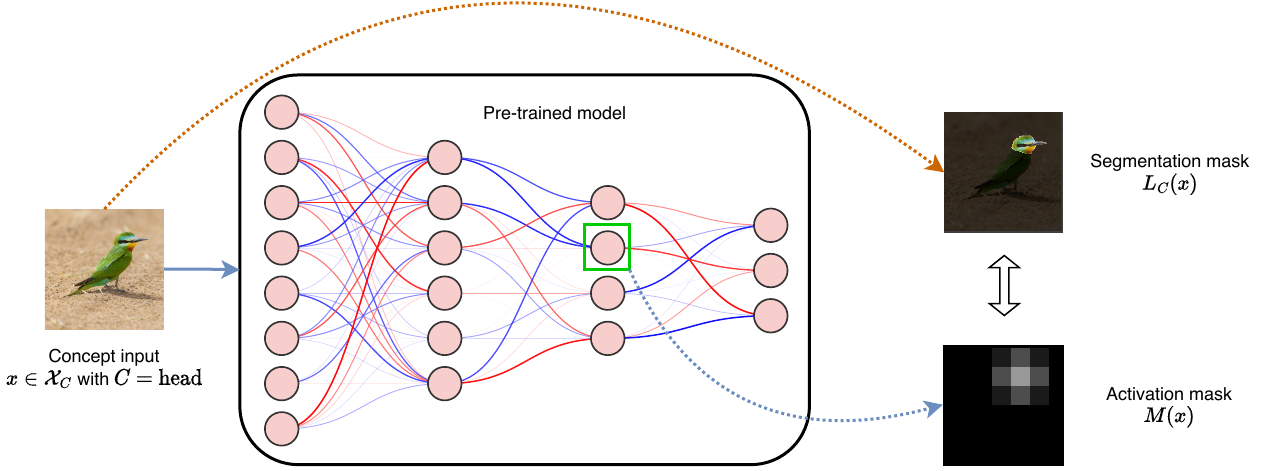}
    \caption{Neuron-level explanation using similarities between concepts and activations. Depicted is the network dissection approach, which compares the segmented concept in the input with the activation mask of a neuron~\cite{bau_network_2017}.}
    \label{fig:dissection}
\end{figure}

In this category, the concept a neuron is representing is explained by comparing the concept with the activations of the neuron when the concept is passed as an input to the model. The \emph{network dissection} approach by Bau et al.~\cite{bau_network_2017} is arguably the most prominent approach in this category, which is mainly applied to computer vision models. In this approach, a set $\mathcal{C}$ of concepts are prepared as well as a set $\mathcal{X}_{C}$ of images for each concept $C \in \mathcal{C}$. Then the activations of a convolutional filter are measured for each input $x \in \mathcal{X}_{C}$. Afterward, the activation map is thresholded to generate a binary activation mask $M(x)$ and scaled up to be compared with the original concept (e.g., concept \texttt{head}) in the binary segmentation mask $L_{C}(x)$ of the input $X$ (e.g., the \texttt{head} segment of an image with a bird). See \autoref{fig:dissection} for an illustration. Then, to measure to which degree concept $C$ is represented by the convolutional filter, the dataset-wide intersection over union metric (IoU) is computed, which is defined as $\text{IoU}(C) = \sum_{x \in \mathcal{X}_{C}}\vert M(x) \cap L_{C}(x)\vert/\sum_{x \in \mathcal{X}_{C}}\vert M(x) \cup L_{C}(x) \vert$. If the IoU value is above a given threshold, then the convolutional filter represents the concept $C$. Several extensions of this approach have been introduced.

Fong et al.~\cite{fong_net2vec_2018} question whether a concept has to be represented by a single convolutional filter alone or whether it can be represented by a linear combination of filters. They show that the latter leads to a better representation of the concept and also suggest to use binary classification for measuring how well filters represent a concept. Complementary to that extension, Mu et al.~\cite{mu_compositional_2020} investigate how to approximate better what a single filter represents. To this end, they assume that a filter can represent a Boolean combination of concepts (e.g., \texttt{(water OR river) AND NOT blue}) and show that this compositional explanation of concepts leads to higher IoU. An intuitive extension of using compositional explanations is using \emph{natural language} explanations. The approach called MILAN by Hernandez et al.~\cite{hernandez_natural_2022} finds such natural language explanations as a sequence $d$ of words that maximizes the pointwise mutual information between $d$ and a set of image regions $E$ that maximally activates the filter, (i.e., $\argmax_{d} \log P(d \mid E) - \log P(d)$). In the approach, the two probabilities $P(d \mid E)$ and $P(d)$ are approximated by an image captioning model and a language model respectively, which are trained on a dataset that the authors curated.

One strong assumption made by the network dissection approach is the availability of a comprehensive set $\mathcal{C}$ of concepts and corresponding labeled images to provide accurate explanations of neurons. This is, however, difficult to obtain in general. Oikarinen et al.~\cite{oikarinen_clip-dissect_2023} tackle this problem with their CLIP-Dissect method, which is based on the CLIP~\cite{radford_learning_2021} vision-language model. (CLIP embeds images and texts in the same vector space, allowing for measuring the similarity between texts and images.) To explain the concept a convolutional filter $k$ is representing, they choose a set $\mathcal{X}_{k}$ of the most highly activating images for filter $k$, then use CLIP to measure the similarity between $\mathcal{X}_{k}$ and each concept $C \in \mathcal{C}$ (here, the concept set $\mathcal{C}$ consists of 20K most common English words), and finally find the best matching concept $C$.

The dissection approach can also be used in generative vision models. Bau et al.~\cite{bau_gan_2018} identify that units of generative adversarial networks~\cite{goodfellow_generative_2014} learn concepts similar to network dissection and that one can intervene on the units and remove specific concepts to change the output image (e.g., removing units representing the concept \texttt{tree} leads to output images with fewer trees in their scenes).


\subsection{Using Causal Relationships between Concepts and Activations}
\label{sec:causality-based}

In this category, the concepts that a neuron is representing are explained by analyzing the causal relationship either (i) between the input concept and the neuron by intervening on the input and measuring the neural activation or (ii) between the neuron and the output concept by intervening on the neural activation and measuring the probability in predicting the concept. This approach is often used for explaining neurons of NLP models \cite{sajjad_neuron-level_2022}, where the types of concepts can be broader (e.g., subject-verb behavior, causal relationship, semantic tags).

The first line of work investigates the influence of a concept in the input on the activation of a neuron by intervening in the input. Kádár et al.~\cite{kadar_representation_2017} find the $n$-grams (i.e., a sequence of $n$ words) that have the largest influence on the activation of a neuron by measuring the change in its activations when a word is removed from the $n$-grams. Na et al.~\cite{na_discovery_2018} first identify $k$ sentences that most highly activate a filter of a CNN-based NLP model. From these $k$ sentences, they extract concepts by breaking down each sentence into a set of consecutive word sequences that form a meaningful chunk. Then they measure the contribution of each concept to the filter's activations by first repeating the concept to create a synthetic sentence of a fixed length (to normalize the input's contribution to the unit across different concepts) and then measuring the mean value of the filter's activations.

\begin{figure}[t]
    \centering%
    \includegraphics[scale=0.7]{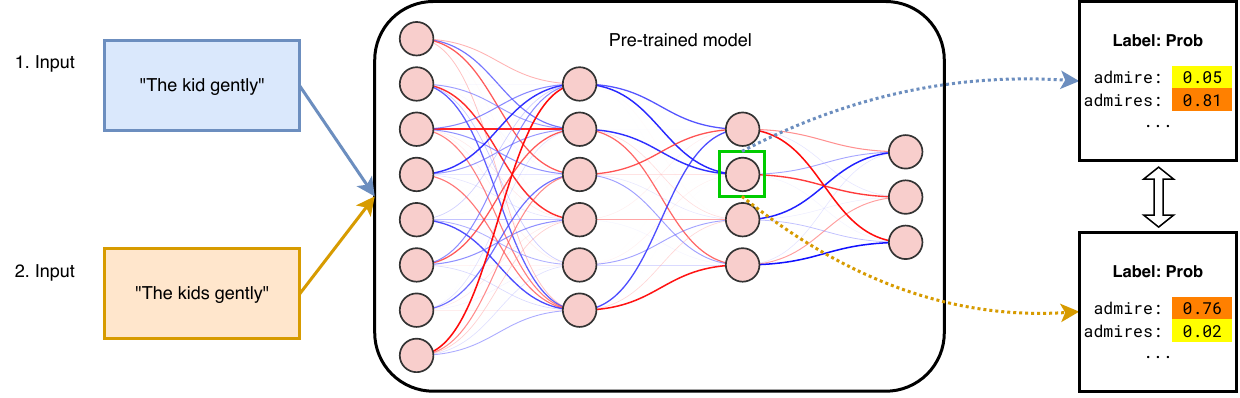}

    
    \caption{Neuron-level explanation using causal relationships between concepts and activations. In causal mediation analysis, the activation of a neuron is modified to the one that the neuron would have output if there was an intervention on the input (the subject in the input sentence was changed from singular to plural). Afterward, the amount of change between the predictions of the correct conjugation of a verb with and without the intervention is measured~\cite{finlayson_causal_2021}.}
    \label{fig:causal-image}
    
\end{figure}

The second line of work investigates the role of a neuron in generating a concept by intervening in the activation of the neuron. Dai et al.~\cite{dai_knowledge_2022} investigate the factual linguistic knowledge of the BERT model~\cite{devlin_bert_2019}, a widely used pre-trained model for text classification, which is pre-trained among other tasks by predicting masked words in a sentence. In this approach, given relational facts with a mask word (e.g. ``Rome is the capital of [MASK]''), each neuron's contribution to predicting the mask is measured using the integrated gradients method~\cite{sundararajan_axiomatic_2017}. To verify the causal role of the neuron that is supposed to represent a concept, the authors also intervene in the neuron's activation (by suppressing or doubling) and measure the change in accuracy in predicting the concept. Finlayson et al.~\cite{finlayson_causal_2021} analyze whether a neuron of a transformer-based language model (e.g., GPT-2~\cite{radford_language_2019}) has acquired the concept of conjugation. The authors determine which neuron contributes most to the conjugation of a verb by using the causal mediation analysis~\cite{vig_investigating_2020}. To this end, they first modify the activation of a neuron to the one that the neuron would have output if there was an intervention on the input (e.g., the subject in the input sentence was changed from singular to plural) and then measure the amount of change between the predictions of the correct conjugation of a verb with and without the intervention (see \autoref{fig:causal-image}). Meng et al.~\cite{meng_locating_2022} also apply causal mediation analysis to GPT-2 to understand which neurons memorize factual knowledge and modify specific facts (e.g., ``The Eiffel Tower is in Paris'' is modified to ``The Eiffel Tower is in Rome''). The data they use consists of triples of the form (\textit{subject}, \textit{relation}, \textit{object}) and the model has to predict the \textit{object} given \textit{subject} and \textit{relation}. They discover that the neurons in the middle layer 
feed-forward modules in GPT-2 are the most relevant for encoding factual information and implementing a weight modifier to change the value of weights and alter the factual knowledge.

\section{Layer-Level Explanations}
\label{sec:explain-activ-layer}

Concepts can also be represented by a whole \emph{layer} as opposed to a neuron or a convolutional filter, as mentioned in the paragraph about the work by Fong et al.~\cite{fong_net2vec_2018} in \autoref{sec:similarity-based}. This can be achieved in a post-hoc manner for a pre-trained model by passing examples of a concept dataset $\mathcal{C}$ to the model and extracting the activations of a specific layer to train a concept classifier. Two approaches are prominent in layer-level explanations: the first is explaining with \emph{concept activation vectors} (CAVs) (see~\autoref{sec:conc-activ-vect}) and the second is \emph{probing} (see~\autoref{sec:probing-expl}). The main difference between the two approaches is that in the case of CAV a linear binary classifier is trained for each concept $C \in \mathcal{C}$, and in probing a multiclass classifier is trained with classification labels that are often related to certain linguistic features (e.g, sentiments, part-of-speech tags). On the other hand, concepts can be baked in a layer, where each concept represents a neuron as was done with localist representations in the early days of neural network research (see~\autoref{sec:conc-bottl-models}).

\subsection{Using Vectors to Explain Concepts: Concept Activation Vectors}
\label{sec:conc-activ-vect}

\begin{figure}[t]
    \centering%
    \includegraphics[scale=0.7]{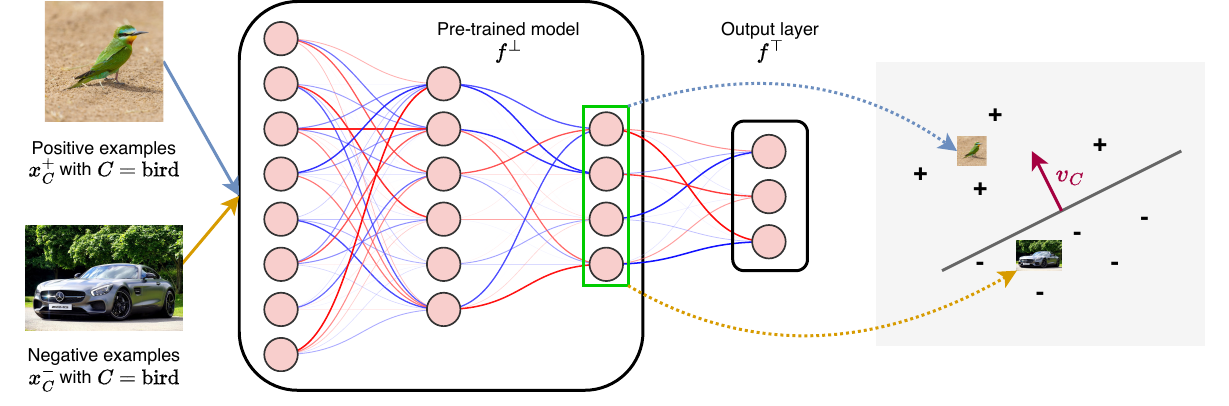}
    \caption{Layer-level explanation using vectors to explain concepts. For each concept $C$ positive examples $x_{C}^{+}$ and negative examples $x_{C}^{-}$ are fed to a pre-trained model to learn the so-called \emph{concept activation vector} (CAV) $v_{C}$ from the corresponding activations of the target layer \cite{kim_interpretability_2018}.}
    \label{fig:cav}
    
\end{figure}

A \emph{concept activation vector} (CAV) introduced by Kim et al.~\cite{kim_interpretability_2018} is a continuous vector that corresponds to a concept represented by a layer of a neural network $f$ 
(see \autoref{fig:cav}). Let $f = f^{\top} \circ f^{\bot}$, where $f^{\bot}: \R^{m} \rightarrow \R^{n}$ is the bottom part of the network whose final convolutional layer $\ell$ is of interest. To identify the existence of a concept $C$ (e.g., the concept \texttt{stripes}) in layer $\ell$, network $f^{\bot}$ is first fed with positive examples $x_{C}^{+}$ that contain concept $C$ and negative examples $x_{C}^{-}$ that do not contain the concept, and then their corresponding activations $f^{\bot}(x_{C}^{+}) \in \R^{n}$ and $f^{\bot}(x_{C}^{-} ) \in \R^{n}$ are collected. Next, a linear classifier is learned that distinguishes activations $f^{\bot}(x_{C}^{+})$ from activations $f^{\bot}(x_{C}^{-})$. The vector normal $v_{C} \in \R^{n}$ to the decision boundary of the classifier is then a CAV of concept $C$. One useful feature of a CAV is that it allows for testing how much an input image $x$ is correlated with a concept $C$ (e.g., an image of a zebra and concept \texttt{stripes}), which is called \emph{testing with CAVs} (TCAV) in \cite{kim_interpretability_2018}. This is accomplished, roughly speaking, by measuring the probability of a concept $C$ having a positive influence on predicting a class label $k \in \{1, \dotsc, K\}$ on a dataset $\mathcal{X}$, i.e., how much moving the latent vector $f^{\bot}(x) \in \R^{n}$ along the direction of $v_{C}$, i.e., $f^{\bot}(x) + \epsilon \cdot v_{C}$, changes the log-probability of label $k$ when it is fed to $f^{\top}$ for all images $x \in \mathcal{X}$ with class label $k$.

CAVs can be used in many different ways. Nejagholi et al.~\cite{nejadgholi_improving_2022} use CAVs to identify sensitivity of abusive language classifiers with respect to implicit types (as opposed to explicit types) of abusive language. Different from the original approach~\cite{kim_interpretability_2018} which obtains CAVs by taking the vector normal to the decision boundary, they obtain CAVs by just averaging over the activations $f^{\bot}(x_{C}^{+})$ for all positive samples $x_{C}^{+}$ to mitigate the impact of the choice of random negative samples $x_{C}^{-}$ on determining the decision boundary. Zhou et al.~\cite{zhou_interpretable_2018} decompose the row vector of the last linear layer for predicting a class label $k$ and represent it as a linear combination of a basis that consists of 
CAVs using only positive weights. Each positive weight then indicates how much of the corresponding concept is involved in predicting class label $k$. Similarly, Abid et al.~\cite{abid_meaningfully_2022} propose an approach that learns a set of CAVs, but for debugging purposes. Given an input image misclassified by a model, a weighted sum of the set of CAVs is computed that leads to correct classification when added to the activations before the last linear layer of the model. In addition to explaining bugs on a conceptual level, this approach allows for identifying spurious correlations in the data.

An issue with the original approach for learning CAVs is that one needs to prepare a set of concept labels and images to learn the CAVs. Ghorbani et al.~\cite{ghorbani_towards_2019} partially tackle this issue by preparing images of the same class and then segmenting them with multiple resolutions. The clusters of resulting segments then form concepts and can be used for TCAV. As corresponding concept labels are missing, the concepts need to be manually inspected. Yeh et al.~\cite{yeh_completeness-aware_2020} circumvent the problem of preparing a concept dataset by training CAVs together with a model on the original image classification dataset. To this end, they compute a vector-valued score, where each value corresponds to a learnable concept and indicates to which degree the concept is present in the receptive field of the convolutional layer (computed by building a scalar product). The score is then passed to a multilayer perceptron (MLP) to perform classification.

\subsection{Using Classifiers to Explain Concepts: Probing}

\label{sec:probing-expl}

\begin{figure}[t]
    \centering%
    \includegraphics[scale=0.7]{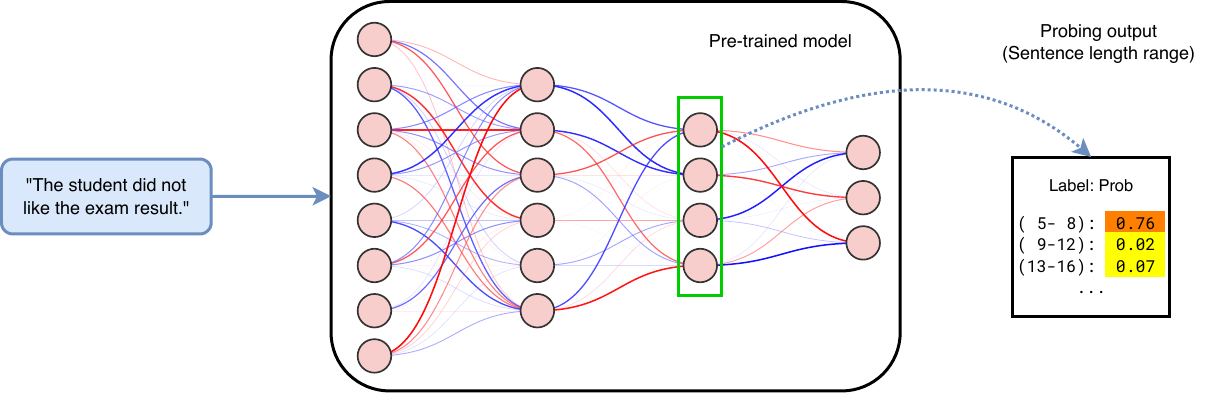}
    \caption{Layer-level explanation using a classifier to explain concepts. In this example, a pre-trained model takes as its input a sentence and a probing classifier is applied to the activation of the highlighted layer to check whether the activation encodes the concept of sentence length~\cite{adi_fine-grained_2016}.}\label{fig:probing}
\end{figure}

Similar to the CAV-based approaches in \autoref{sec:conc-activ-vect}, \emph{probing} uses a classifier to explain concepts. However, instead of training a binary linear classifier for each concept $C \in \mathcal{C}$ to measure the existence of the concept in the activation of a layer, probing uses a classifier for multiclass classifications with labels that often represent linguistic features in NLP (e.g., sentiments, part-of-speech tags). For example, given sentences as inputs to a pre-trained NLP model (e.g., BERT~\cite{devlin_bert_2019}), probing allows for evaluating how well the sentence embeddings of the model capture certain syntactic and semantic information, such as the length or the tense of the sentence~\cite{ettinger_probing_2016,adi_fine-grained_2016,conneau_what_2018} (see \autoref{fig:probing}).

Probing, which is designed as a layer-level explanation method, can also be combined with a neuron-level explanation method (see~\autoref{sec:expl-activ-neur}) by applying the probing classifier only to neurons that are relevant for the classification~\cite{durrani_analyzing_2020}. Finding such neurons can be accomplished by applying the elastic-net regularization to the classifier, which constrains both the L1- and the L2-norm of the classifier weights.

The concepts learned by such probing classifiers can be combined with a knowledge base to provide richer explanations. Ribiero and Leite~\cite{ribeiro_aligning_2021} use identified concepts as evidence to draw conclusions from a set of axioms in a knowledge base (e.g., given an axiom $\mathtt{LongFreightTrain} \leftarrow \mathtt{LongTrain} \wedge \mathtt{FreightTrain}$ in the knowledge base, identifying both antecedent concepts $\mathtt{LongTrain}$ and $\mathtt{FreightTrain}$ in the activations explains the presence of the consequence $\mathtt{LongFreightTrain}$ in the input). However, one cannot always assume the presence of a knowledge base for a given task. Ferreira et al.~\cite{ferreira_looking_2022} weaken this assumption by learning the underlying theory from the identified concept using an induction framework.
    


Since the probing classifier is trained independently from the pre-trained model, it is pointed out that the pre-trained model does not necessarily leverage the same features that the classifier uses for predicting a given concept, i.e., what the probing classifier detects can be merely a correlation between the activation and the concept~\cite{belinkov_probing_2022,amini_naturalistic_2022}.

\subsection{Using Localist Representations: Concept Bottleneck Models}
\label{sec:conc-bottl-models}

\begin{figure}[t]
    \centering %
    \includegraphics[scale=0.7]{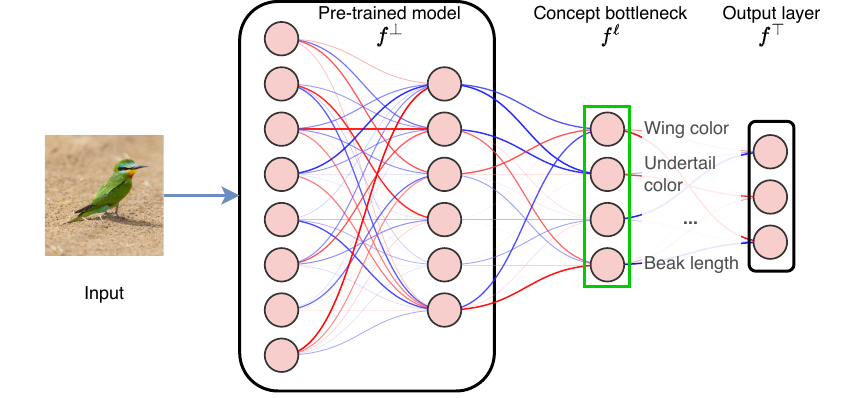}
    \caption{Layer-level explanation using the concept bottleneck model approach~\cite{koh_concept_2020}. Each neuron in the concept bottleneck $f^{\ell}$ corresponds to a unique concept (e.g., \texttt{wing color}).}
    \label{fig:cbm}
\end{figure}

Different from the neuron-based approach in \autoref{sec:expl-activ-neur}, where concepts are learned in a post-hoc manner, in a \emph{concept bottleneck model} (CBM)~\cite{koh_concept_2020}, each concept is represented by a unique neuron in the bottleneck layer $f^{\ell}$ of a model $f$ (see \autoref{fig:cbm}), which is a reminiscence of localist representations~\cite{page_connectionist_2000}. This layer provides information about the existence or the strengths of each concept in the input. The output of the bottleneck layer is then used by a classifier or regressor $f^{\top}$ for the prediction, which allows for explaining what concept led to the given prediction. Often, the bottom part $f^{\bot}$ of a pre-trained model is used for initializing the layers before the concept bottleneck $f^{\ell}$ and $f^{\ell}$ is a linear layer that maps the features from $f^{\bot}$ to concepts. Therefore, a CBM is $f = f^{\top} \circ f^{\ell} \circ f^{\bot}$, where $\circ$ stands for composition. To train the concept bottleneck $f^{\ell}$, the training data has to include concept labels in addition to task labels.

One of the main limitations of CBMs is the need for the aforementioned concept labels, which might not be available for specific tasks. Several recent approaches overcome this limitation~\cite{yuksekgonul_post-hoc_2023,oikarinen_label-free_2023,yang_language_2023}. The main idea behind these approaches is using an external resource to obtain a set $\mathcal{C}$ of concepts relevant to the task. This external resource could be a knowledge base such as ConceptNet~\cite{yuksekgonul_post-hoc_2023}, or the 20K common English words~\cite{oikarinen_label-free_2023}, or a language model like GPT-3~\cite{yang_language_2023}. After obtaining concept set $\mathcal{C}$, each concept word $C \in \mathcal{C}$ is embedded as a vector $v_{C}$ by means of the CLIP vision-language model (cf.~\autoref{sec:similarity-based}) such that vector $v_{C}$ can be used for computing the strength of concept $C$ for a given input $x \in \mathcal{X}$, e.g., by measuring the cosine similarity between $v_{C}$ and the embedding $f^{\bot}(x)$. Finally, the presence of concepts in the concept bottleneck layers allows for inducing logical explanations, e.g., Ciravegna et al.~\cite{ciravegna_logic_2022} induce explanations in disjunctive normal form (DNF) from concept activations and predicted labels, which is similar to logic-based explanation approaches~\cite{ribeiro_aligning_2021,ferreira_looking_2022} in \autoref{sec:probing-expl}.

\section{Conclusion}
\label{sec:conclusion}
In this survey, we have reviewed recent methods for explaining concepts in neural networks. We have covered different approaches that range from analyzing individual neurons to learning classifiers for a whole layer. As witnessed by the increasing number of recent papers, this is an active research area and a lot is still to be discovered, for example, empirically comparing or integrating different approaches.\footnote{For example, concept bottleneck models in \autoref{sec:conc-bottl-models} and concept activation vectors in \autoref{sec:conc-activ-vect} have been combined by Yuksekgonul et al.~\cite{yuksekgonul_post-hoc_2023}.} With the progress of concept extraction from neural networks, integrating the learned neural concepts with symbolic representations---also known as \emph{neuro-symbolic integration}---is receiving (again) increasing attention~\cite{ribeiro_aligning_2021,ferreira_looking_2022,ciravegna_logic_2022,dalal_understanding_2023,barbiero_interpretable_2023,lecue_role_2020}. In the near future, we expect tighter integration between neural models and symbolic rules via concept representations to make the models more transparent and easier to control. In conclusion, this line of research is still very active and in development, providing ample opportunities for new forms of integration in neuro-symbolic AI.

\section{Acknowledgement}
The authors gratefully acknowledge support from the DFG (CML, MoReSpace, LeCAREbot), BMWK (SIDIMO, VERIKAS), and the European Commission (TRAIL, TERAIS). We would like to thank Cornelius Weber for valuable comments on this paper.


\bibliographystyle{ios1} %

\bibliography{main}

\end{document}